\documentclass[10pt,twocolumn,letterpaper]{article}

\usepackage{cvpr}              % To produce the CAMERA-READY version
\usepackage{graphicx}
\usepackage{amsmath}
\usepackage{amssymb}
\usepackage{booktabs}
\usepackage{multirow}

\usepackage[pagebackref,breaklinks,colorlinks]{hyperref}

\usepackage[capitalize]{cleveref}
\crefname{section}{Sec.}{Secs.}
\Crefname{section}{Section}{Sections}
\Crefname{table}{Table}{Tables}
\crefname{table}{Tab.}{Tabs.}

\def\cvprPaperID{*****} % *** Enter the CVPR Paper ID here
\def\confName{CVPR}
\def\confYear{2022}

\begin{document}

\title{One-stage Action Detection Transformer}

\author{Lijun Li, Li'an Zhuo, Bang Zhang\\
Alibaba Group\\
{\tt\small \{shenfei.llj, lianzhuo.zla, zhangbang.zb\}@alibaba{-}inc.com}
}
\maketitle

%%%%%%%%% ABSTRACT
\begin{abstract}
   In this work, we introduce our solution to the EPIC-KITCHENS-100 2022 Action Detection challenge. One-stage Action Detection Transformer (OADT) is proposed to model the temporal connection of video segments. With the help of OADT, both the category and time boundary can be recognized simultaneously. After ensembling multiple OADT models trained from different features, our model can reach 21.28\% action mAP and ranks the 1st on the test-set of the Action detection challenge.
\end{abstract}

%%%%%%%%% BODY TEXT
\section{Introduction}
\label{sec:intro}
With the explosion of video contents, video understanding has gained lots of interest from computer vision researchers~\cite{ji20123d,karpathy2014large, wang2018temporal,lijuncaption,lijunattack}. In this field, action related tasks form the basis of video understanding. Compared with traditional action recognition~\cite{simonyan2014two, lijunaction,wu2019long}, action detection not only recognizes action classes, but also detects the temporal boundaries simultaneously. Although only solve one another task, it is much difficult to distinguish the boundary since the action interval is ambiguous. In order to solve the action detection task, most traditional works firstly generate action proposals sorted by confidence score, then use another separate module to classify the proposals. With the great success of transformer in vision, a few works start to insert transformer into the action detection pipeline~\cite{actionformer,xu2021long}. We follow similar pipeline and propose a one-stage network OADT for action detection.
%-------------------------------------------------------------------------

\section{Our Approach}
\label{sec:approach}
\begin{figure*}[ht]
  \includegraphics[width=\linewidth,height=7cm]{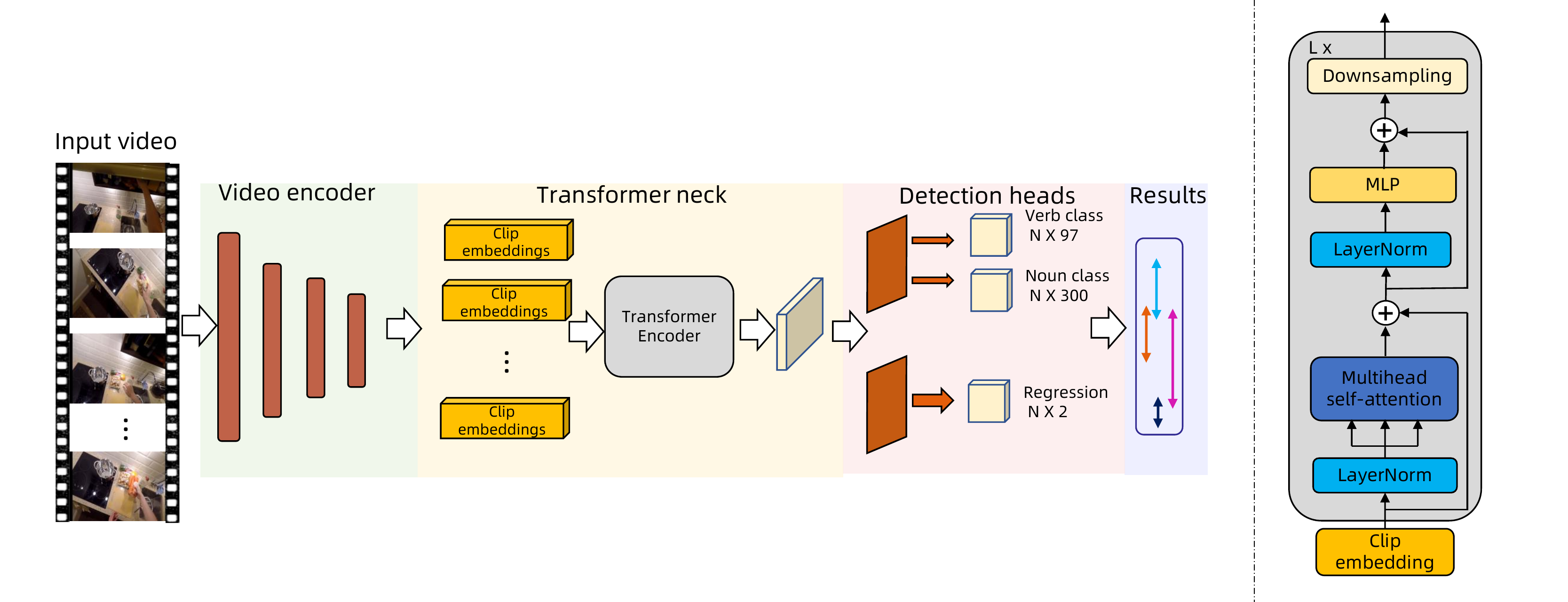}
  \caption{Overview of our proposed OADT. It is composed of three parts: video encoder which extracts clip-level features from untrimmed videos, transformer neck that takes in the clip embeddings and performs self-attention and detection heads which classify the clips and regress the time boundary.}
  \label{fig:pipeline}
\end{figure*}

The overall structure is showed in Fig.~\ref{fig:pipeline}. The network is composed of three parts: video encoder, transformer neck and detection heads. In the following, we will describe each part in details.

\subsection{Video Encoders.}
\label{sec:encoder}
Limited by the device memory, the raw video cannot be directly fed to the network. Therefore, the clip-level features are extracted from the untrimmed video using the video encoders. The video encoders are adapted from action recognition without the classification head. In this work, five superior action recognition methods are implemented.

\paragraph{Omnivore~\cite{girdhar2022omnivore}.} Omnivore is based on the swin-transformer, which leverages the flexibility of transformer-based architectures and is trained jointly on classification tasks from different modalities. For action recognition, the videos are converted into spatio-temporal tubes, and then these tubes are projected into embeddings using the linear layer.

\paragraph{MVit~\cite{fan2021multiscale}.} Multiscale Vision Transformers create a multiscale pyramid of features on the vision transformer, which hierarchically expands the feature complexity while reducing visual resolution.

\paragraph{Motionformer~\cite{patrick2021keeping}.} Motionformer introduces the trajectory attention that aggregates information along implicitly determined motion paths on the video transformer.

\paragraph{Slowfast~\cite{feichtenhofer2019slowfast}.} SlowFast proposes a two-pathway architecture for video recognition. A slow pathway with a low frame rate is designed to capture spatial semantics. In contrast, a fast pathway, operating at high temporal resolution, is responsible for dealing with rapid motion.

\paragraph{TimeSformer~\cite{gberta_2021_ICML}.} TimeSformer is a transformer-based approach built exclusively on self-attention over space and time, where temporal attention and spatial attention are separately applied within each block.

\subsection{Transformer Neck}
The transformer neck is composed of a sequence of transformer~\cite{vaswani2017attention} layers. It takes in the clip embeddings obtained by the video encoder and performs self-attention. As is shown in the right of Fig.~\ref{fig:pipeline}, the basic transformer layer includes the layer norm (LN) operations~\cite{ba2016layer}, multi-head self-attention (MHSA), residual connections~\cite{he2016deep}, multi-layer perceptron (MLP) and the downsampling operation. Furthermore, a feature pyramid with different temporal resolutions is created to capture the various temporal range of actions.

\subsection{Detection Heads}
Different from the two-stage approaches that generate the segment proposals firstly, The detection heads solve the action classification and segment regression in a synchronous manner. The detection heads predict $N$ results directly, where $N$ is the predefined maximum of the proposals. 
For the regression head, the segments including the begin and end time are predicted by the several full-connection layers.
For the classification head, verb and noun are also predicted by the full-connection layers separately for corresponding proposals, and then both are combined into action classification using the simple operations, i.e., addition or multiplication. 
The focal losses~\cite{lin2017focal} are employed on optimizing verb, noun and action classification, and the 1D IOU losses are used for segment regression. 
% Furthermore, We introduce SoftNMS to combine the action classification and the segment regression into the final detection results.

\begin{table*}[!t]
  \small
  \centering
  \begin{tabular}{{lccccccc}}
    \toprule
    \multirow{2}{*}{Team} & \multirow{2}{*}{Label}&\multicolumn{6}{c}{Test mAP(\%)}\\ 
    % \cline{3-8}
    & &@0.1 &@0.2 &@0.3 &@0.4&@0.5 &Avg\\
    \midrule
    \multirow{3}{*}{richard61}  &Verb   &22.78 &21.68 &20.14 &18.34 &15.54 &19.69\\
                                &Noun   &19.33 &17.98 &16.55 &14.69 &12.28 &16.17\\
                                &Action &14.33 &13.63 &12.80 &11.53 & 9.93 &12.44\\
    % \hline
    \cline{2-8}
    \multirow{3}{*}{Bristol-MaVi}   &Verb   &25.33 &23.99 &21.91 &19.61 &17.08 &21.58\\
                                    &Noun   &18.99 &17.87 &16.41 &14.43 &11.36 &15.81\\
                                    &Action &14.71 &13.98 &12.86 &11.56 & 9.85 &12.59\\
    % \hline
    \cline{2-8}
    \multirow{3}{*}{CTC-AI}     &Verb   &22.62 &21.73 &20.68 &17.74 &15.16 &19.58\\
                                &Noun   &20.65 &19.58 &18.34 &16.18 &12.88 &17.52\\
                                &Action &16.68 &16.11 &15.15 &13.59 &11.66 &14.64\\
    % \hline
    \cline{2-8}
    \multirow{3}{*}{Alibaba-MMAI-Research} &Verb   &22.77 &22.01 &19.63 &17.81 &14.65 &19.37 \\
                                           &Noun   &26.44 &24.55 &22.30 &19.82 &16.25 &21.87 \\
                                           &Action &18.76 &17.73 &16.26 &14.91 &12.87 &16.11 \\
    % \hline
    \cline{2-8}
    \multirow{3}{*}{4Paradigm-UWMadison-NJU}   &Verb   &27.11 &26.07 &24.38 &21.96 &18.59 &23.62 \\
                                               &Noun   &28.71 &27.27 &25.19 &22.33 &\textbf{18.82} &24.47 \\
                                               &Action &23.73 &22.87 &21.36 &19.53 &\textbf{16.86} &20.87 \\
    \midrule
    \multirow{3}{*}{Ours}   &Verb   &\textbf{30.67} &\textbf{29.40} &\textbf{26.81} &\textbf{24.34} &\textbf{20.51} &\textbf{26.35} \\
                            &Noun   &\textbf{30.96} &\textbf{29.36} &\textbf{26.78} &\textbf{23.27} &18.80 &\textbf{25.83} \\
                            &Action &\textbf{24.57} &\textbf{23.50} &\textbf{21.94} &\textbf{19.65} &16.74 &\textbf{21.28} \\
    \bottomrule
  \end{tabular}
  \caption{Final results on EPIC-KITCHENS-100 test set.}
  \label{tab:compare}

\end{table*}

\section{Experiments}
Epic-KITCHENS-100~\cite{damen2022rescaling} is a large-scale egocentric action dataset. The dataset is very challenging because it contains various kinds of verb and noun classes from fine-grained action videos which capture all daily activities in the kitchen.
\subsection{Experimental Details}
 In this challenge, we employ the video classification methods and pretrain them on Kinetics600~\cite{carreira2018short} dataset firstly. Then they are finetuned on EPIC-KITCHENS-100 dataset for action recognition. After finetuning, clip-level features are generated with sliding windows. For each sliding window, the time interval is 32 frames and the temporal stride is 16 frames. In the training stage of action detection, the model is trained for 27 epochs and the input resolution is 456$\times$ 256. AdamW~\cite{loshchilov2017decoupled} optimizer is used with weight decay of 0.0005. The batch size is 2 and the learning rate is set to 0.0001 with the cosine scheduler. We generate the action labels by combining verb and noun predictions. The corresponding time intervals are obtained from the regression head. In inference, Soft-NMS~\cite{bodla2017soft} is used for post-processing to suppress redundant action segments.

\begin{table}
%   \footnotesize
  \scriptsize
  \centering
  \begin{tabular}{{lcccccc}}
    \toprule
    \multirow{2}{*}{Video encoder}  & \multicolumn{6}{c}{Val mAP(\%) for Action}\\ 
    % \cline{2-7}
    &@0.1 &@0.2 &@0.3 &@0.4&@0.5 &Avg\\
    \midrule
    TimeSformer~\cite{gberta_2021_ICML}        &20.47 &19.75 &18.69 &17.02 &14.82 &18.15\\
    SlowFast~\cite{feichtenhofer2019slowfast}  &21.01 &20.15 &19.02 &17.66 &15.13 &18.59\\
    MVit~\cite{fan2021multiscale}              &22.41 &21.44 &20.16 &18.50 &16.10 &19.72\\
    Motionformer~\cite{patrick2021keeping}     &22.99 &22.08 &20.64 &18.73 &16.09 &20.11\\
    Omnivore~\cite{girdhar2022omnivore}        &25.38 &24.50 &23.09 &21.18 &18.72 &22.57\\
    \midrule
    Ensemble      &\textbf{27.19} &\textbf{26.23} &\textbf{24.38} &\textbf{22.47} &\textbf{19.82} &\textbf{24.02}\\
    \bottomrule
  \end{tabular}
  \caption{Detection results on EPIC-KITCHENS-100 validation set.}
  \label{tab:val}

\end{table}

\paragraph{Evaluation metrics.} Mean Average Precision (mAP) is used to evaluate verbs, nouns and actions at different temporal IOU thresholds as well as average mAP. In EPIC-KITCHENS-100 dataset, temporal IOU thresholds range from 0.1 to 0.5 with a step of 0.1. We follow the official split of training, validation and test. For test submission, our model is first trained on the training\&validation set and then test on the test set.

\paragraph{Ensemble models.}
In order to further boost our performance, we apply the five action recognition methods mentioned in Section~\ref{sec:encoder} as video encoder separately and train each OADT model. To make full use of different models, we ensemble OADT model trained from different features as our final model.

% \begin{table}
%   \centering
%   \begin{tabular}{{l|ccc}}
%     \toprule
%     \multirow{2}{*}{Feature}  & \multicolumn{3}{c}{mAP(\%) for val}\\ 
%     &Verb &Noun &Action\\
%     \midrule
%     Omnivore      & -& -& 22.57\\
%     Motionformer  & -& -& 20.11\\
%     MultiscaleVit & -& -& 19.72\\
%     SlowFast      & -& -& 18.53\\
%     TimeSformer   & -& -& 18.15\\
%     \midrule
%     Ensemble      & -& -& 23.97\\
%     \bottomrule
%   \end{tabular}
%   \caption{Detection results on EPIC-KITCHENS-100 validation set.}
%   \label{tab:val}

% \end{table}

% \begin{table*}[t]
%   \centering
%   \begin{tabular}{{l|c|cccccc|cccccc}}
%     \toprule
%     \multirow{2}{*}{Feature} & \multirow{2}{*}{Method} & \multicolumn{6}{c}{mAP(\%) for val} & \multicolumn{6}{c}{mAP(\%) for test}\\ 
%     & &@0.1 &@0.2 &@0.3 &@0.4&@0.5 &Avg &@0.1 &@0.2 &@0.3 &@0.4 &@0.5 &Avg\\
%     \midrule
%     Omnivore & OADT \\
%     MultiscaleVit & OADT\\
%     %Motionformer & OADT &22.99 &22.08 &20.64 &18.73 &16.09 \\
%     SlowFast & OADT\\
%     TimeSformer & OADT\\
%     Ensemble & - &- &- &- &- &- &- &24.57 &23.50 &21.94 &19.65 &16.74 &21.28
%     \bottomrule
%   \end{tabular}
%   \caption{Final results for action on EPIC-KITCHENS-100 test set.}
%   \label{tab:action}

% \end{table*}

\paragraph{Results.}
Tab.~\ref{tab:val} shows our results on validation set. From the table, we can see that OADT using Omnivore as the video encoder performs best. While Motionformer and MVit perform slightly worse and are roughly 2\% lower. In the end, we ensemble all the five models and can reach 24.02\% which is about 1.45\% higher than the single best model in action mAP.
In Tab.~\ref{tab:compare}, we compare our result to existing state-of-the-art results on the test set. Our solution can get 21.28\% mAP which is 5\% higher than the winning solution of last year. We outperform prior work especially on verb class by a large margin of +3\%.
%-------------------------------------------------------------------------

\section{Conclusion}
\label{sec:Conclusion}
We present our OADT model used in the EPIC-KITCHENS-100 2022 Action Detection challenge. Our model is a one-stage transformer-based architecture composed of the video encoder, transformer block and multiple heads for classification and regression. After ensembling five models, our model can reach the state-of-the-art result of 21.28\% average mAP on the EPIC-KITCHENS-100 test set.
%-------------------------------------------------------------------------

\paragraph{Acknowledgement}
{We would like to thank the whole EPIC-KITCHENS team for hosting such a great challenge.}

%%%%%%%%% REFERENCES
{\small
\bibliographystyle{ieee_fullname}
\bibliography{egbib}
}

\end{document}